\documentclass[conference]{IEEEtran}
\IEEEoverridecommandlockouts
\usepackage{cite}
\usepackage{amsmath,amssymb,amsfonts}
\usepackage{algorithmic}
\usepackage{graphicx}
\usepackage{textcomp}
\usepackage{xcolor}
\usepackage{multirow}
\usepackage{adjustbox}
\usepackage{comment}
\usepackage{soul}
\usepackage{booktabs}
\usepackage{hyperref}

\newcommand{\eg}{\textit{e}.\textit{g}.}

\def\BibTeX{{\rm B\kern-.05em{\sc i\kern-.025em b}\kern-.08em
    T\kern-.1667em\lower.7ex\hbox{E}\kern-.125emX}}
\begin{document}

\title{Empirical Study of Mix-based Data Augmentation Methods in Physiological Time Series Data}

\author{\IEEEauthorblockN{Peikun Guo}
\IEEEauthorblockA{\textit{Rice University} \\
Houston, USA \\
pg34@rice.edu}
\and
\IEEEauthorblockN{Huiyuan Yang}
\IEEEauthorblockA{\textit{Rice University} \\
Houston, USA \\
hy48@rice.edu}
\and
\IEEEauthorblockN{Akane Sano}
\IEEEauthorblockA{\textit{Rice University} \\
Houston, USA \\
akane.sano@rice.edu}
}

\maketitle

\begin{abstract}
Data augmentation is a common practice to help generalization in the procedure of deep model training. In the context of physiological time series classification, previous research has primarily focused on label-invariant data augmentation methods. However, another class of augmentation techniques (\textit{i.e., Mixup}) that  emerged in the computer vision field has yet to be fully explored in the time series domain.
In this study, we systematically review the mix-based augmentations, including mixup, cutmix, and manifold mixup, on six physiological datasets, evaluating their performance across different sensory data and classification tasks. 
Our results demonstrate that the three mix-based augmentations can consistently improve the performance on the six datasets. More importantly, the improvement does not rely on expert knowledge or extensive parameter tuning.
Lastly, we provide an overview of the unique properties of the mix-based augmentation methods and highlight the potential benefits of using the mix-based augmentation in physiological time series data. Our code and results are available at \url{https://github.com/comp-well-org/Mix-Augmentation-for-Physiological-Time-Series-Classification}.
\end{abstract}


\section{Introduction}

Data augmentation is a crucial regularization technique for deep neural network models, as it serves to inform the network of potential variations in the input data during the training stage while preserving the integrity of the labels. This technique has been shown to improve network generalization,\cite{krizhevsky2017imagenet} by not only artificially increasing the size of the dataset but also imparting inductive bias through the encoding of information related to data invariances.

Traditional data augmentation techniques aim to increase the statistical support of the training data distribution by utilizing human knowledge and adding additional virtual samples from the vicinity distribution of training samples. This approach has been shown to improve generalization, as demonstrated in previous literature. Such data augmentations have been employed actively and effectively in computer vision \cite{simard1998transformation, shorten2019survey, krizhevsky2017imagenet, zhang2017mixup} and speech recognition and synthesis \cite{jaitly2013vocal, ko2017study, park2019specaugment}. The selection of specific augmentation methods remains a challenging task, as it is often based on heuristics and is highly dependent on the dataset, task, and even model architecture \cite{ho2019population}. 
However, unlike in other domains, time series, particularly physiological data, do not follow a straightforward rule for label-invariant transformation. Methods such as jittering, rotation, scaling, permutation, magnitude warping, time warping, window slicing, and window warping have been shown to have unstable performance across datasets and tasks, or require human understandings of the data \cite{huiyuan, wen2020time, iwana2021empirical}. Two significant limitations of traditional transformation-based augmentations on physiological time series data are that: (1) certain transformations can be detrimental to the integrity of the physiological signal, and (2) the majority of traditional augmentations are data-dependent, lacking generalization and consistency across different datasets and tasks.

An alternative approach is represented by mixup regularization \cite{zhang2017mixup}, which is based on the assumption that linear interpolations of feature vectors should lead to linear interpolations of the associated targets. Despite its simplicity, mixup has been shown to be effective across different domains (computer vision \cite{zhang2017mixup, verma2019manifold, carratino2020mixup} and speech \cite{meng2021mixspeech, jia2022emotion, xu2018mixup}) and different tasks. For time series classification tasks, previous studies also employed mix-based augmentations to enhance model representation and generalization \cite{antoni2022automatic, han2021towards, han2023improving}. However, none of the previous works provide a thorough empirical study of mix-based augmentations across various types of physiological times series, regarding both the quantitative gain in the metrics and the benefits of feature representation.

\begin{figure*}[htp!]
\begin{center}
\centerline{\includegraphics[width=0.85\linewidth]{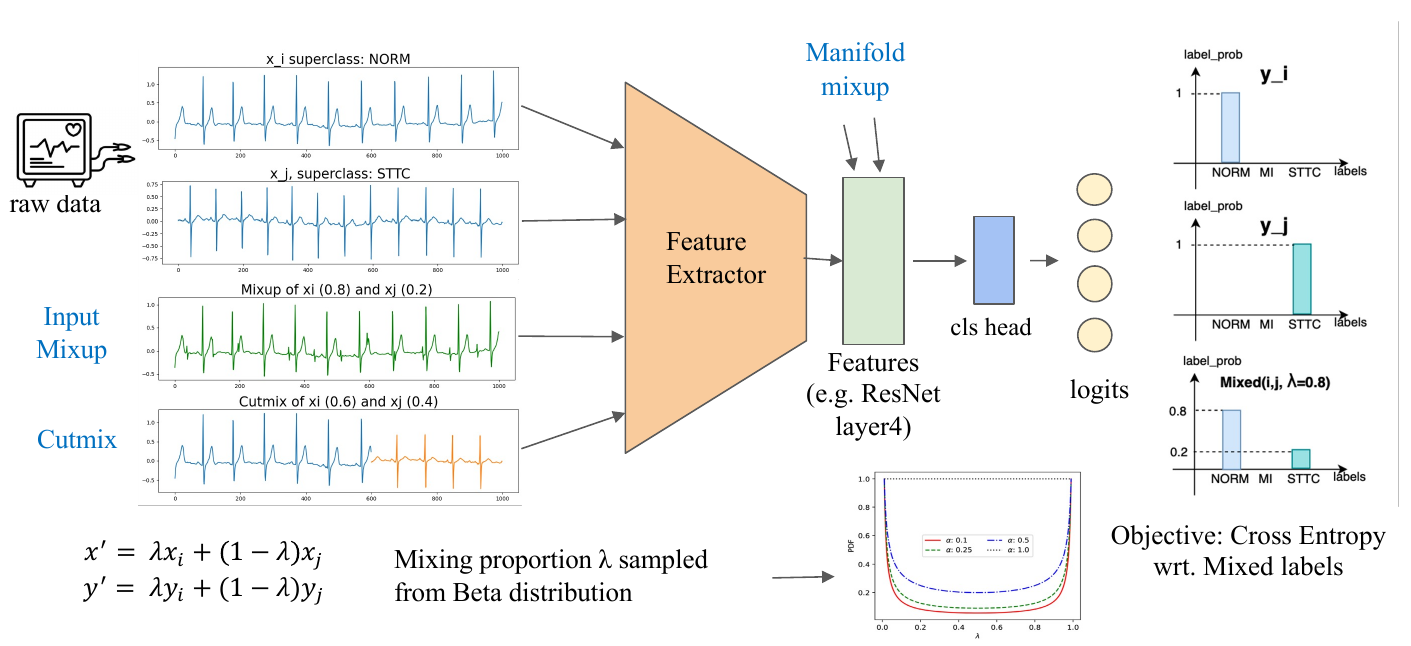}}
\caption{The overview of mix-based augmentation procedure for physiological time series classification. From left to right: two sequences $x_i, x_j$ are shown as raw input signals. For mixup or cutmix, in each training epoch, virtual samples are created from the mini-batches; for manifold mixup, the linear combination is applied at the feature map level. For any of the three mix-based augmentations, the labels for the virtual samples are also mixed with the corresponding weights, shown on the right. NORM, MI, and STTC are representative classes of the PTB-XL dataset. See details in Figure \ref{ptb-meta}.}
\label{overview}
\end{center}
\vskip -0.2in
\end{figure*}


This study aims to evaluate the efficacy of mix-based data augmentation in the context of time series classification. Our evaluation compares mix-based augmentation against traditional data augmentation techniques, as classified in a previous survey, focusing on basic label-invariant time-domain transformations commonly used in time series classification. The baseline augmentations evaluated include jittering, rotation, scaling, permutation, magnitude warping, time warping, window slicing, and window warping. The unique benefits of mix-based augmentation are evaluated, and the contributions of this paper can be summarized as follows:

\begin{itemize}
    \item 
    We present an empirical study of three mixup-based data augmentation methods (i.e., mixup, cutmix, and manifold mixup) in the context of time series classification. We provide detailed formulations of these methods and evaluate their performances on six physiological and biobehavioral datasets.
    \item
    Our experiments reveal two significant distinctions between mixup-based augmentations and traditional data transformations. First, mixup-based methods do not require human expert priors, making them more practical in various applications. Second, mixup-based methods consistently achieve higher or comparable performance compared to traditional methods.
\end{itemize}

\section{Related works}\label{related-works}

\subsection{Traditional label-invariant time series data augmentation}\label{sec2.1}

For physiological time series data, the time domain transforms manipulate the original time series directly, as compared to more advanced augmentations using generative approaches. In this paper, we focus the evaluations on the following traditional augmentations: jittering, rotation, scaling, permutation, window warping, and window slicing. 


Jittering refers to the injection of Gaussian or more sophisticated noise patterns, such as spikes and slope-like trends, into the raw signals. The schemes are introduced in \cite{wen2019time}. Rotation for time series is achieved by multiplying the signal by a random rotation matrix. As it is not as suitable for time series data as in the image domain, one commonly used special case is flipping (changing the sign of the original time series). Permutation is a transformation that randomly shuffles segments of a time series. This operation does not preserve the sequential information of the original data. Window slicing or cropping, introduced in \cite{cui2016multi}, randomly extracts continuous slices from the original samples. The window warping method is uniquely applicable to time series. It randomly selects a time interval, then upsamples or downsamples the segment, while keeping the rest of the time ranges unaltered. Window warping changes the total length of the original signal, therefore it is usually used along with window cropping. For time series classification tasks, all the above-mentioned augmentations do not change the labels of the altered training samples.

The effectiveness of these augmentations has been previously investigated in the literature. An empirical study about biobehavioral time series data augmentation \cite{huiyuan} concludes that, while some augmentations are beneficial for biobehavioral classification tasks, their effectiveness varies across different datasets and model architectures. This finding agrees with \cite{iwana2021empirical}, which also highlights the inconsistency of transformation-based augmentations across different non-physiological time series datasets.

Based on the analysis of the literature, we claim that two significant limitations exist when applying traditional transformation-based augmentations to physiological time series data:

\begin{itemize}
\item
    Certain transformations can be detrimental to the integrity of the physiological signals. For instance, rotation, permutation, and warping can be more harmful to ECG signals, as ECG beats possess relatively fixed patterns, such as the order, width, and intensity of the wave components.
\item
The majority of traditional augmentations are data-dependent, meaning that they require expert prior knowledge or multiple trials to select an appropriate transformation for a specific problem, due to the diverse properties of biobehavioral time series. 
\end{itemize}

\subsection{Mixup: Vicinal Risk Minimization}\label{vicinal-risk-minimization-with-mixup-in-computer-vision}

Mixup is a data augmentation technique introduced by \cite{zhang2017mixup} to train neural networks by constructing virtual training examples using convex combinations of pairs of examples and their labels. 


The methodology behind Mixup, as described in this paper, is rooted in Vicinal Risk Minimization (VRM), which diverges from the conventional Empirical Risk Minimization (ERM) by drawing examples from a vicinity distribution of the training examples. This aims to enlarge the support for the training distribution. As stated in Section \ref{related-works}, prior knowledge has been traditionally required for the identification of the vicinity or neighborhood. However, Mixup provides a more practical and data-agnostic alternative, as it does not necessitate domain expertise. Despite its simplicity, Mixup presents the following distinct advantages:

\begin{enumerate}
\def\labelenumi{\arabic{enumi}.}
\item
  \textbf{Regularization}: The linear relationship established by mixup transformations between data augmentation and the supervision signal results in a strong regularization of the model's state, leading to improved performance.
\item
  \textbf{Generalization}: The authors of previous studies have reported improvements in generalization error for state-of-the-art models trained on ImageNet, CIFAR, speech, and tabular datasets when using mixup. Theoretical analysis \cite{carratino2020mixup} suggests that the soft targets of mixup virtual samples aid in model generalization in a manner similar to label smoothing and knowledge distillation. Interpolation/extrapolation of nearest neighbors in feature space can also improve generalization \cite{devries2017dataset}.
  
\end{enumerate}

Since mixup was proposed, many incremental works have emerged and demonstrated improvements with different focuses, such as cutmix \cite{yun2019cutmix}, manifold mixup \cite{verma2019manifold}, MixMatch \cite{berthelot2019mixmatch}, and AlignMix \cite{venkataramanan2022alignmixup}. In this project, we choose to evaluate mixup, cutmix, and manifold mixup, as they have not been investigated in the context of time series classification in literature.

\subsection{Mixup for physiological time series data}\label{mixup-TS}
In the domain of physiological time series classification, mixup has been employed to enhance generalization during training as demonstrated in prior studies.  \cite{antoni2022automatic, han2021towards} employ mixup for better generalization in ECG classification task, in the training batches of CNN models. \cite{han2023improving} states that mixup improves the generalization performance of the ECG classification model regardless of leads and evaluation metrics. However, these studies lack a thorough examination of mixup's mechanism and reasons for performance improvement, and none of them have conducted ablation studies about mixup. Furthermore, although the 1D variant of vanilla Mixup \cite{zhang2017mixup} has been utilized in previous studies, the empirical results for cutmix \cite{yun2019cutmix} and manifold mixup \cite{verma2019manifold} are currently lacking. In \cite{yang2022robust}, the performance of mixup and cutmix were evaluated on the UEAMTSC dataset \cite{bagnall2018uea} using InceptionTime \cite{ismail2020inceptiontime} as the baseline model. However, the subsets of UEAMTSC in that study were of small scale and not strictly comprised of time series data (\eg image contours). This paper, on the other hand, aims to investigate the effectiveness of the mix-based methods on various physiological time series datasets, utilizing a higher capacity residual network structure.

\begin{figure}[htp!]
\vskip -0.1in
\begin{center}
\centerline{\includegraphics[width=1.0\columnwidth]{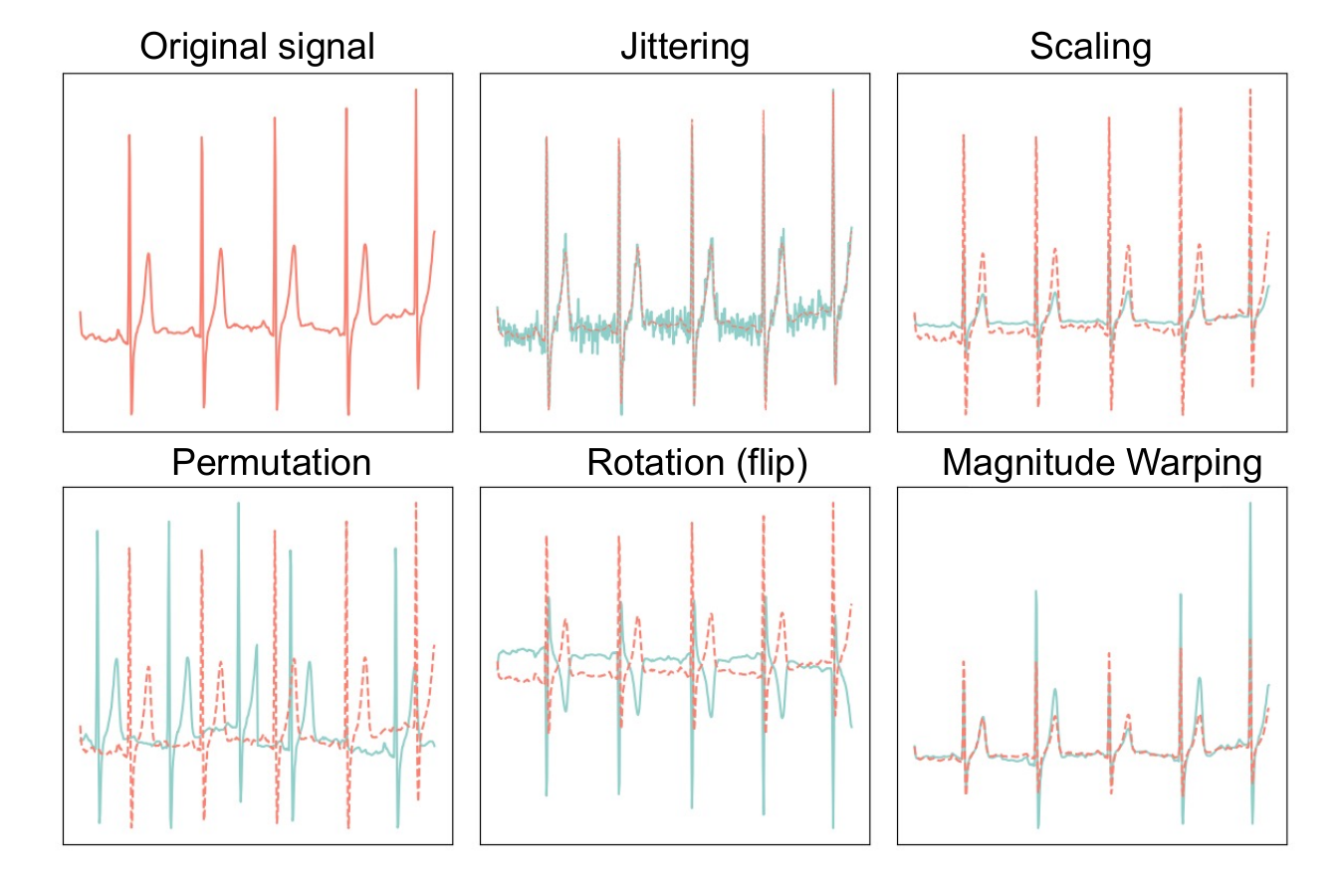}}
\caption{Illustrations of traditional time series data augmentations. Orange: original signal. Green: augmented signals.}
\label{traditional_da}
\end{center}
\vskip -0.2in
\end{figure}

\section{Augmentation methods}\label{augmentation-methods}
In this section, we provide a formal introduction and implementation of mix-based augmentations. 
Figure \ref{overview} shows the overall paradigm of how the mix-based augmentation is applied during the training process. The intrinsic difference between mix-based and traditional augmentations will be discussed in section \ref{results-and-discussion}. The technical details for the implementation of the augmentations can be found in section \ref{experimental-setup}.

\subsection{Cutout}\label{cutout}

In this study, the Cutout augmentation serves as a benchmark for comparison against mix-based data augmentation methods. Cutout, originally introduced in the computer vision field to address occlusion issues, involves the removal of contiguous sections of data. To evaluate its effectiveness in the context of time series classification, a single-item transformation, similar to traditional label-invariant regularizations, is applied. Specifically, a random contiguous section of a time series is replaced with zeros, which can be considered a dropout (zero-masking) operation at the input layer. The size of the random time segment is fixed, while the starting index of the interval is randomly drawn from a uniform distribution, and applied to all channels of a single data point.

\subsection{Mixup}\label{mixup}

The time series mix-based augmentations in this study leverage multiple data samples from one training minibatch to generate virtual data points. We examine three commonly utilized configurations of mix-based augmentations from the computer vision domain for time series classification problems, namely mixup \cite{zhang2017mixup}, cutmix \cite{yun2019cutmix}, and manifold mixup (layer mixup, \cite{verma2019manifold}).

The mixup augmentation blends random pairs of time series from the training data. Let $(x, y)$ denote a time series data instance, where $x \in \mathbb{R}^{L \times C}$, with $L$ representing the length of the sequence and $C$ denoting the number of channels, and $y \in \mathbb{R}^{K}$ being the class label with $K$ classes. The mixing ratio randomly drawn from a Beta distribution is denoted as $\lambda$. Given two samples $(x_{i}, y_{i})$ and $(x_{j}, y_{j})$, the mixup augmentation generates virtual training examples through the following formulation:

\begin{align}
\tilde{x} &= \lambda x_i + (1-\lambda) x_j \\
\tilde{y} &= \lambda y_i + (1-\lambda) y_j 
\end{align}


Mixup extends the training distribution by incorporating the
the assumption that linear interpolations of feature vectors should lead to linear interpolations of the associated targets \cite{zhang2017mixup}. Based on the formulation, the label for a mixup virtual sample is also a mixture of two original one-hot labels weighted by \(\lambda\). The potential effect and benefit of the soft target will also be discussed in section \ref{profiling-mix-based-augmentations-on-ptb-xl}.

The mixing ratio, $\lambda$, in the mix-based data augmentation approach is sampled from a Beta distribution. The value of $\lambda$ close to 0 or 1 results in the created virtual time series being more similar to one of the raw data points, whereas a value of $\lambda$ close to 0.5 results in a more blended representation of the raw data points. For physiological time series data, it is desirable to have virtual time series that are similar to one of the raw data points, as these signals contain delicate features, such as the intensity of R peaks in ECG signals, which could be easily destroyed with random mixing. 

Despite its simplicity, in the computer vision domain, mixup has allowed consistently superior performance in the CIFAR-10, CIFAR-100, and ImageNet image classification datasets\cite{zhang2017mixup}. As we will show in the results section, mixup can also improve classification metrics in time series classification problems, along with other desirable features.

\subsection{Cutmix}\label{cutmix}


The image augmentation technique Cutmix, proposed in \cite{yun2019cutmix}, shares similarities with the technique Mixup. The authors claim a key advantage of Cutmix is its ability to prevent the occurrence of ambiguous components in the generated samples caused by mixing, such as blurred image regions \cite{yun2019cutmix}. For time series cutmix, we select a random time segment from a pair of multivariate time series, then the values of the pair of time series within the segment are exchanged across all channels. The length of the segment is also determined randomly through the mixing ratio $\lambda$ drawn from a Beta distribution. 

\subsection{Manifold Mixup}\label{manifold-mixup-layer-mixup}

Manifold Mixup, presented in \cite{verma2019manifold}, demonstrates consistently superior performance across various computer vision tasks when compared to the original input-data-mixup approach. Unlike the original mixup, manifold mixup trains neural networks on linear combinations of hidden representations of training samples. The literature suggests that higher-level representations obtained from intermediate layers of the neural network feature extractor are low-dimensional, therefore, linear interpolations of hidden representations should cover meaningful regions of the feature space. In this paper, we implement Layer Mixup on layer 4 of the ResNet, prior to pooling and the classification head. The labels are also mixed in the same way as mixup and cutmix.

\section{Experiments}\label{experiments}
\subsection{Datasets}\label{datasets}

We conduct experiments on six biomedical time series datasets, encompassing diverse data types and varying sizes. The datasets include two ECG datasets, PTB-XL for cardiac condition classification and Apnea-ECG for sleep apnea detection, two EEG datasets, Sleep-EDF for sleep stage recognition and MMIDB for sleep movement detection, and two IMU datasets, PAMAP2 and UCI-HAR for human activity recognition. Table \ref{dataset} provides a summary of the datasets. Note that in column "Periodic", the IMU datasets are tagged as "motion", because IMU data may contain periodic patterns when the recorded activity contains periodic motion (\eg walking).

\begin{table*}[t]
\caption{Dataset Summary}
\label{dataset}
\begin{center}
\begin{small}
\begin{sc}
\begin{tabular}{lcccccccr}
\hline
Dataset & Category & \# channels & \# classes & Sample length & Periodic & \# samples \\
\hline
PTB-XL    & ECG &  12 & 5 & 1000 & Yes & 17962   \\
Apnea-ECG & ECG &  1  & 2 & 6000 & Yes & 34243 \\
Sleep-EDFE  & EEG  &  1 & 5 & 3000 & No & 42308\\
MMIDB-EEG    & EEG & 64 & 2 & 640 & No & 4635 \\
PAMAP2      & IMU  & 52 & 12 & 1000 & (motion) & 5452 \\
UCI-HAR     & IMU & 9 & 6 & 128 & (motion) & 10299 \\
\hline
\end{tabular}
\end{sc}
\end{small}
\end{center}
\vskip -0.1in
\end{table*}

\subsubsection{PTB-XL}\label{ptb-xl}

The PTB-XL dataset \cite{wagner2020ptb} is an ECG database of 12-lead recordings, containing 44 diagnostic statements grouped into 5 superclasses (normal, conduction disturbance, myocardial infarction, hypertrophy, and ST-T change). In this study, we formulate a five-class cardiac abnormality classification problem. The data is divided into 10 balanced folds, with the first 8 used for training, the 9th for validation, and the 10th for testing. The data we use is sampled at 100Hz. The same stratification process is applied to the other datasets if no validation/test set is provided.


\begin{figure}[ht]
\begin{center}
\centerline{\includegraphics[width=0.9\columnwidth]{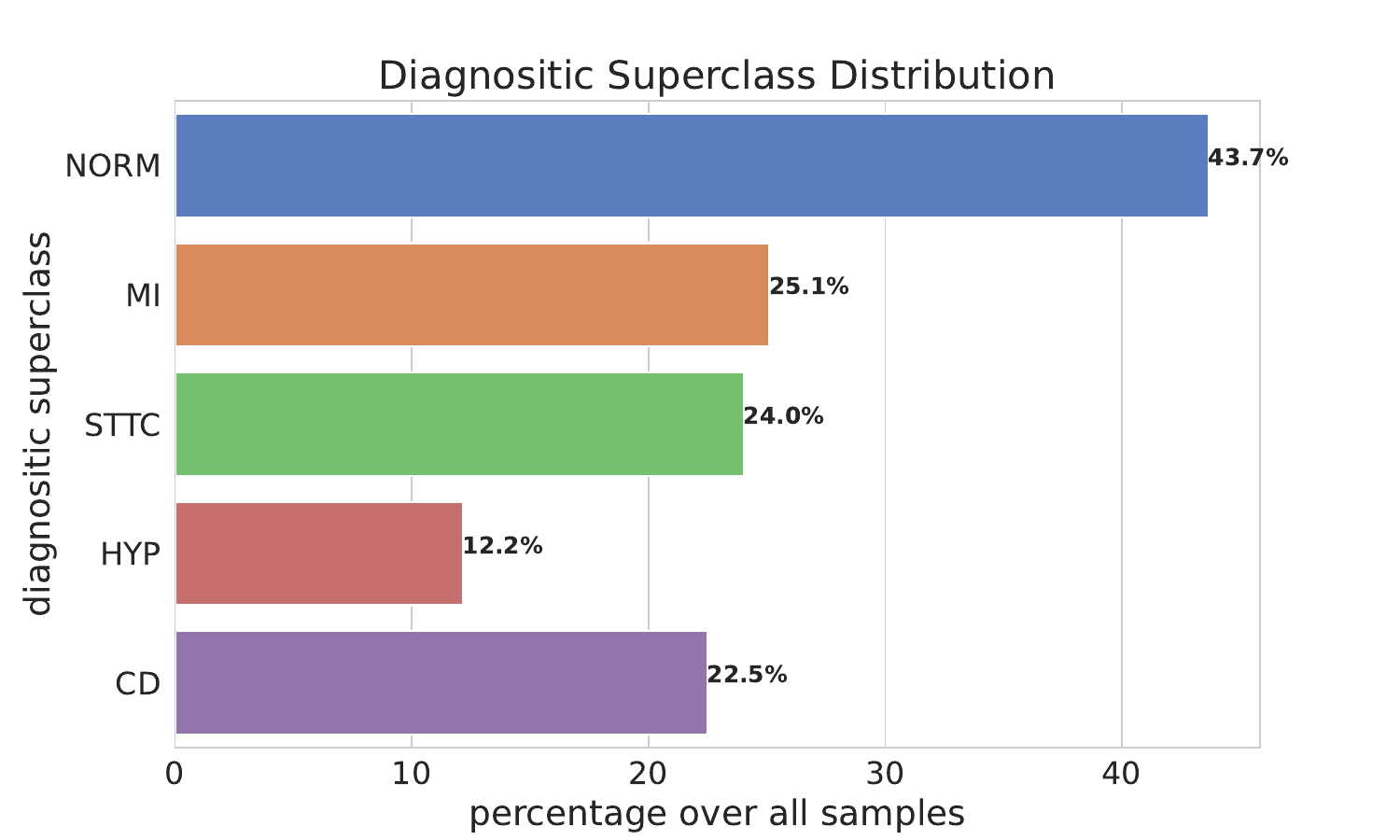}}
\centerline{\includegraphics[width=0.75\columnwidth]{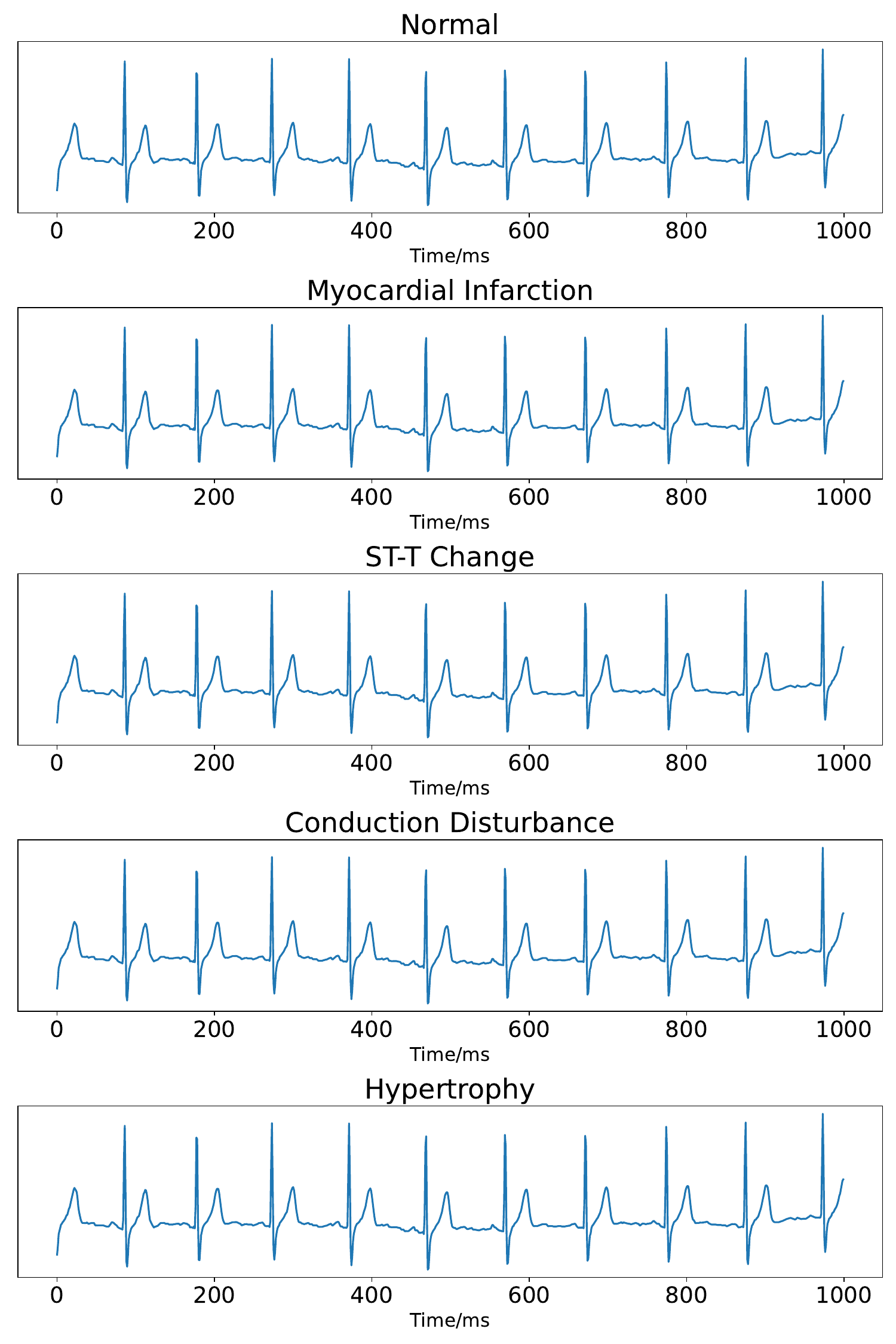}}
\caption{PTB-XL dataset overview. Top: The five-superclass distribution of PTB-XL (NORM, MI, STTC, CD, HYP). In this paper, we only train with single-label samples. Bottom: Example waveforms of PTB-XL classes (lead IV), the x-axis denotes time in ms, and the y-axis is the normalized voltage. }
\label{ptb-meta}
\end{center}
\vskip -0.3in
\end{figure}

The dataset is highly class-imbalanced as shown in Figure \ref{ptb-meta}, with over half of the samples labeled normal and the least-represented class (HYP, hypertrophy in left ventricle) only constituting 3.3\% of the data. This challenge is addressed in Section \ref{results-and-discussion} through in-batch resampling and mix-based augmentations.

\subsubsection{Apnea-ECG}\label{apnea-ecg}


The Apnea-ECG dataset examines the connection between sleep apnea symptoms and heart activity in humans \cite{penzel2000apnea}, as monitored through ECG. This dataset has 70 records, sampled at 100 Hz, with 35 records designated for training and the remaining 35 for testing. Each record is 7-10 hours in length and includes a continuous ECG signal along with per-minute apnea annotations indicating the presence or absence of sleep apnea. We segmented the ECG recordings into 60-second frames at 100 Hz, resulting in 17233 samples for the training set and 17010 samples for the test set.

\subsubsection{Sleep-EDFE}\label{sleep-edfe}


The Sleep-EDFE dataset is sourced from the publicly accessible Sleep European Data Format (EDF) database \cite{kemp2018sleep} on Physionet \cite{moody2001physionet}. This database contains full-night PSG sleep records that include two-channel EEG (Fpz-Cz and Pz-Oz), a horizontal EOG, and EMG signal records, along with corresponding hypnograms (sleep stage annotations). The EEG signals have a sampling frequency of 100 Hz, and they are divided into 30-second epochs and normalized to have zero mean and unit standard deviation.

\subsubsection{MMIDB-EEG}\label{mmidb-eeg}


The EEGMMIDB (EEG Motor Movement/Imagery Database) from PhysioNet is collected using the BCI200 EEG system4 \cite{schalk2004bci2000}. It records 64 channels of brain signals at a sampling rate of 160 Hz, totaling over 1500 recordings of 1-2 minutes each. Subjects are instructed to wear the EEG device and sit in front of a computer screen, performing specific typing tasks in response to on-screen prompts.

\subsubsection{PAMAP2}\label{pamap2}


The PAMAP2 dataset \cite{reiss2012pamap} comprises recordings from 9 participants who were asked to perform 12 daily activities, including household tasks and various exercises (e.g., Nordic walking, playing soccer). Data from accelerometers, gyroscopes, magnetometers, temperature sensors, and heart rate monitors are recorded from inertial measurement units placed on the hand, chest, and ankle over 10 hours, resulting in a 52-dimensional dataset.

\subsubsection{UCI-HAR}\label{uci-har}


The UCI-HAR dataset \cite{reyes2016transition} was collected from 30 volunteers aged 19 to 48 years. Participants were instructed to engage in six basic activities, which included three static postures (standing, sitting, lying) and three dynamic activities (walking, walking downstairs, walking upstairs). 3-axial accelerometer and gyroscope signals were recorded at a constant rate of 50 Hz. The data was collected using smartphones carried by the participants.

\subsection{Experimental Setup}\label{experimental-setup}
\subsubsection{Network architecture}\label{network-architecture}

In all experiments, we use a 1D-CNN-based ResNet-18 \cite{he2016deep} as the backbone. This model has convolutions with a kernel size of 3, and stride 2. The blocks in the ResNet architecture have convolutional layers with 32, 64, 128, and 256 channels respectively. The output after the final block is average pooled in the temporal dimension, and then a linear layer is applied to predict the probability of the positive class. Details of the ResNet-18 structure are summarized in Table.\ref{ResNet18-Model-structure}.


\begin{table}[]
\caption{Structure of the ResNet-18 backbone. $B$: batch size; $L$: length of sequence; $C$: number of channels.}
\label{ResNet18-Model-structure}
\centering
\begin{adjustbox}{max width=0.48\textwidth}
\begin{tabular}{cccc}
\hline Layer Name & Input Shape & Output Shape & Parameter \\
\hline Reshape Conv1d & {$[\mathrm{B}, \mathrm{L}, \mathrm{C}]$} & {$[\mathrm{B}, \mathrm{C}, \mathrm{L}]$} & $-$ \\
Layer1 & {$[\mathrm{C}, \mathrm{L}]$} & {$[\mathrm{B}, 64, \mathrm{~L}]$} & $1 \mathrm{x} 3,64, \max$ pool \\
Layer2 & {$[\mathrm{B}, 64, \mathrm{~L} / 2]$} & {$[\mathrm{B}, 64, \mathrm{~L} / 2]$} & {$[[1 \mathrm{x} 3,64],[1 \mathrm{x} 3,64]] \mathrm{x} 2$} \\
Layer3 & {$[\mathrm{B}, 128, \mathrm{~L} / 4]$} & {$[\mathrm{B}, 128, \mathrm{~L} / 4]$} & {$[[1 \mathrm{x} 3,128],[1 \mathrm{x} 3,128]] \mathrm{x} 2$} \\
Layer4 & {$[\mathrm{B}, 256, \mathrm{~L} / 8]$} & {$[\mathrm{B}, 512, \mathrm{~L} / 16]$} & {$[[1 \mathrm{x} 3,512],[1 \mathrm{x} 3,256]] \mathrm{x} 2$} \\
Average pool & {$[\mathrm{B}, 512, \mathrm{~L} / 16]$} & {$[\mathrm{B}, 512,1]$} & $-$ \\
FC & {$[\mathrm{B}, 512]$} & {$[\mathrm{B}$, Classes $]$} & {$[512$, Classes $]$} \\
\hline
\end{tabular}
\end{adjustbox}
\end{table}

For the manifold mixup experiments, we take the output of layer 4 of ResNet18 as the input for mixing. 
We also perform t-SNE visualization on the features extracted from Layer 4 as validation for the quality of class representation.

\subsubsection{Augmentation Implementation}\label{implementation-of-augmentations}

Following the previous empirical studies for mix-based regularization: mixup \cite{zhang2017mixup} and MixMatch \cite{berthelot2019mixmatch}, we use \(\alpha=0.4\) and \(\alpha=0.75\) for mix-related hyperparameters in our experiments. For cutmix, the ratio of the random segment length to the signal length is set to 0.2, following \cite{yun2019cutmix}. For the baseline augmentations, the hyperparameters, such as the intensity of scaling and jittering, are manually chosen following \cite{huiyuan}.





Following \cite{zhang2017mixup}, the implementation of the training step is based on the mini-batches sampled by a data loader. For each minibatch, random shuffling is applied, and the mixing operations can all be performed in a vectorized manner, incurring minimal computation overhead. In our PTB-XL profiling experiments, the mean increase in the processing time of each mini-batch of size 128 is less than 0.001 second. We also report results for baselines (denoted as vanilla) that does not use any data augmentation.

\subsubsection{Optimization}\label{optimization-details}

We use the AdamW optimizer with learning rate 0.001 for all datasets. For the profiling experiments on PTB-XL dataset, we also tested with Adam optimizer and 0.005 learning rate. We use a step decay learning rate scheduler with step size 5, and a decay rate of 0.9 across all experiments. Training takes 50 epochs, which was observed to be sufficient for convergence in all datasets.

\subsubsection{Computing resource}: All model training was performed on a single NVIDIA GTX 2080Ti GPU. 

\section{Results and discussion}\label{results-and-discussion}

\subsection{Quantitative performance of mix-base augmentations}\label{quantitative-performance-of-mix-based-augmentations}

\begin{table*}[t]
\caption{Classification accuracies for mix-based augmentation methods and three traditional augmentations are reported on six datasets. The warp column refers to window warping. Bold numbers indicate the best performance.}
\label{results_table}
\begin{center}
\begin{small}
\begin{sc}
\begin{tabular}{lccccccccr}
\hline
Dataset & baseline & Mixup & Cutmix & Layer mixup & Cutout & Jitter & Scale & Permute & Warp\\
\hline
PTB-XL    & 77.94 & 78.91 & \textbf{79.64} & 78.97 & 77.61 & 77.60 & 78.03 & 76.57 & 75.67 \\
Apnea-ECG & 78.10 & 74.73 & 78.30 & \textbf{79.69} & 78.27 & 76.44 & 70.64 & 77.26 & 76.34\\
Sleep-EDFE    & 83.40 & 84.27 & 84.40 & 84.48 & 83.84 & 81.55 & 83.40 & \textbf{85.01} & 83.98\\
MMIDB-EEG    & 78.64  & 79.83  & 77.99 & \textbf{80.58} & 80.04  & 79.07 & 80.15 & 69.58 & 77.99\\
PAMAP2      & 93.62  & 95.31  & \textbf{95.83} & 93.88  & 86.98 & 86.85 & 88.18 & 85.82 & 87.59\\
UCI-HAR     & 92.77 & 93.62 & \textbf{94.26} & 93.11 & 93.58  & 87.95 & 88.63 & 91.89 & 92.94\\
\hline
\end{tabular}
\end{sc}
\end{small}
\end{center}
\vskip -0.1in
\end{table*}

Experiments were conducted on six datasets of diverse categories with a ResNet18-1D backbone. The performance of vanilla (no augmentation), cutout, and three mix-based augmentations are presented in Table \ref{results_table}. We summarize our results as follows.

I).\textbf{The mix-based data augmentations can achieve superior accuracy in comparison to traditional augmentation methods.} All three mix-based augmentation techniques were found to outperform the baseline (no augmentation) in 16 out of 18 experimental trials, with only two exceptions (mixup for Apnea-ECG and cutmix for MMIDB-EEG). Furthermore, among the six datasets examined, the majority of the highest  accuracy results were achieved through the use of cutmix and layer mixup.   

Overall, the mix-based augmentations outperform the baselines, but the performance of augmentation methods varies across different datasets. This is likely due to the unique characteristics of each dataset and the strengths of each augmentation method. For instance, datasets containing complex temporal patterns or high levels of noise may benefit from the use of certain mix-based augmentation methods that are particularly effective at enhancing classification accuracy. The results in Table \ref{results_table} also show that more comprehensive mixup schemes (cutmix, manifold mixup) help yield better accuracy.


II).\textbf{The mix-based augmentations deliver robust performance, and the accuracy gain is steady.}
In addition to the quantitative performance advantages, these techniques are notable for their low dependency on expert knowledge and parameter tuning. Across all 18 mix-based experimental trials (excluding cutout), no significant reduction in  accuracy was observed in comparison to the baseline. In contrast, traditional data transformation techniques can often result in a drastic decrease in accuracy if not implemented appropriately. For example, applying scaling in Apnea-ECG resulted in a 7.5\% reduction in accuracy, permutation in MMIDB-EEG resulted in a 9.0\% reduction, and jittering in UCI-HAR resulted in a 4.8\% reduction. This finding implies that traditional transformations can undermine crucial features in physiological signals, such as wave intensity in ECG data and temporal correlations in EEG data. As a result, these augmentations can generate virtual samples that deviate from the actual data distribution, potentially compromising the generalization performance of the model.

Note that in the experiments, we compared the mix-based augmentations against some individual traditional augmentations. However, to fully harness the potential of data augmentation, it is compatible to apply mix-based augmentations in conjunction with one or more traditional augmentations.

\subsection{Profiling mix-based augmentations on PTB-XL}\label{profiling-mix-based-augmentations-on-ptb-xl}

\begin{figure}[ht]
\vskip -0.1in
\begin{center}
\centerline{\includegraphics[width=1.0\columnwidth]{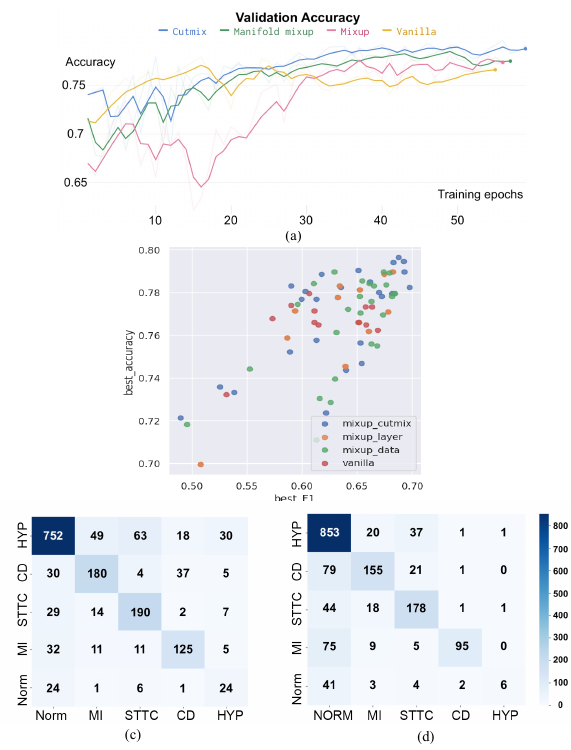}}
\caption{(a): The PTB-XL validation accuracy computed after each training epoch, with baseline and variations of mixup. (b): The scatter plot of validation accuracy again F1 score for all 80 profiling experiments on PTB-XL dataset, different colors annotate different augmentations. (c) and (d): The confusion matrix of cutmix training, with or without a balanced sampler, respectively. The y-axis represents the true labels, and the x-axis is the predicted labels.}
\label{ptb_plot}
\end{center}
\end{figure}

The PTB-XL dataset \cite{wagner2020ptb} is a well-studied ECG dataset for cardiac condition classification, with the current SOTA accuracy of recognizing five classes being less than 80\% \cite{huiyuan}. To evaluate the effectiveness of mix-based data augmentation methods, extensive profiling experiments were conducted on the PTB-XL dataset, using over 80 combinations of settings and hyperparameters (Section \ref{optimization-details}). The scatter plot of the best validation accuracy vs the best F1 score is shown in Figure \ref{ptb_plot}(a). Identical combinations of learning rates, optimizers, etc. were used for each of the four augmentation setups (vanilla, input mixup, cutmix, layer mixup). The scatter plot illustrates that mix-based augmentations produce the best results, with all top-performing results (1st to 28th) obtained from mix-based augmentations, supporting the conclusions drawn in Section \ref{quantitative-performance-of-mix-based-augmentations}.

The PTB-XL dataset presents an imbalanced class distribution for the single-label data samples. To mitigate the high false negative rate for the minority class, a batch class-balanced data sampler was utilized in the data loader during the training process. The effect of the class-balanced sampler on the model's performance is shown in Figure \ref{ptb_plot} (c) and (d), by comparing the confusion matrices with and without the balanced sampling.  As observed in the bottom row of Figure \ref{ptb_plot}, which corresponds to the data samples of the minority class hypertrophy (HYP), the model without the balanced sampler was prone to producing the most false negative (NORM) predictions for the HYP cases. However, by incorporating both the class-balanced sampler and cutmix, virtual samples containing the information of the minority class were generated during the training, reducing the false negative rate. Additionally, the recall of the other three cardiac condition classes also received similar improvement as shown in the plot.

\subsection{Feature representations}\label{feature-representations}

\begin{figure}[ht]
\begin{center}
\centerline{\includegraphics[width=\columnwidth]{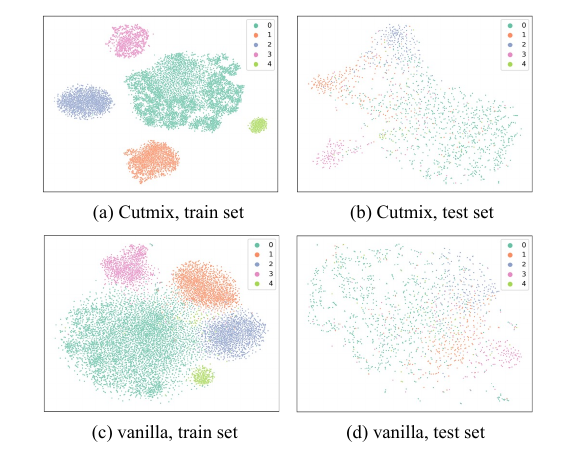}}
\caption{The t-SNE visualizations of cutmix and vanilla settings after training on PTB-XL. In both experiments, a balanced data sampler is used during training to increase the performance of minority classes.}
\label{tsne_plot}
\end{center}
\vskip -0.2in
\end{figure}

The advantages of mix-based augmentations, or vicinal risk minimization, include the provision of more distinguishable representations for different classes. We present the results of t-SNE dimensional reduction of two models trained with cutmix and a baseline, on the training and test sets of the PTB-XL dataset. The feature vectors were calculated using layer 4 of ResNet18. The visualizations of the training set (Figure \ref{tsne_plot}(a) and (c)) indicate that, compared to the baseline, cutmix gives more discriminative representations between classes. The projections of the test set (Figure \ref{tsne_plot}(b) and (d)) demonstrate that the classes are more distinguishable with mix-based augmentation and balanced sampling. 

\section{Conclusion and Future Work}\label{future-work}

Inspired by the success of mixup in other domains, we investigate mixup and its variants, cutmix and manifold mixup for physiological for the time series classification task. This paper empirically shows that mix-based data augmentation techniques can achieve superior accuracy in comparison to traditional augmentation methods in the context of time series classification. In the experiments, the majority of the highest accuracy results were achieved through the use of cutmix and layer mixup, and these augmentations were found to deliver robust performance with a steady accuracy gain across various physiological and biobehavioral datasets. The low dependency on expert knowledge and parameter tuning, in addition to the quantitative performance advantages, makes mix-based augmentations more practical and effective in various applications. These findings highlight the effectiveness of mix-based, dataset-agnostic augmentations and the importance of appropriately choosing traditional data transformations, as they can compromise the generalization performance of the model.

We plan to explore the combination of mix-based augmentation and traditional time series augmentations, as the effectiveness of well-composited transformations has been shown in previous studies \cite{iwana2021empirical}. Furthermore, we aim to extend the applicability of mix-based augmentation to the frequency domain, following the success of such an approach in acoustic data classification tasks \cite{kim2021specmix}.

\bibliographystyle{IEEEtran}  
\bibliography{main}

\end{document}